\documentclass[10pt,twocolumn,letterpaper]{article}

\usepackage[pagenumbers]{cvpr} 

\usepackage{graphicx}
\usepackage{amsmath}
\usepackage{amssymb}
\usepackage{bm}
\usepackage{booktabs}
\usepackage{cases}
\usepackage{floatrow}
\usepackage{multirow}
\usepackage{pifont}  
\usepackage{xspace}
\usepackage{array,colortbl,xcolor}
\usepackage{placeins}
\usepackage{stfloats}
\usepackage[accsupp]{axessibility}  

\usepackage[pagebackref,breaklinks,colorlinks]{hyperref}

\usepackage[capitalize]{cleveref}
\crefname{section}{Sec.}{Secs.}
\Crefname{section}{Section}{Sections}
\Crefname{table}{Table}{Tables}
\crefname{table}{Tab.}{Tabs.}

\newcommand{\cmark}{\ding{51}}
\newcommand{\xmark}{\ding{55}}
\newcommand{\kl}[2]{\operatorname{KL}({#1} \; || \; {#2})}

\newcommand{\mc}[1]{\mathcal{#1}}
\newcommand{\Cb}{\mc{C}^b}
\newcommand{\Cn}{\mc{C}^n}
\newcommand{\Nb}{{|\Cb|}}
\newcommand{\Nn}{{|\Cn|}}
\newcommand{\pp}{\mathbf{p}}

\newcommand{\qq}{\mathbf{q}}
\newcommand{\xx}{\mathbf{x}}
\newcommand{\yy}{\mathbf{y}}

\newcommand{\mi}[2]{\text{I}(#1; #2)}
\newcommand{\ent}[1]{\text{H}(#1)}
\newcommand{\support}{\mathbb{S}}

\newcommand{\thetabf}{\boldsymbol{\theta}}
\newcommand{\thetabase}{\thetabf_b}
\newcommand{\thetanovel}{\thetabf_n}

\newcommand{\prior}{\boldsymbol{\Pi}}
\newcommand{\ourmethod}{\text{DIaM}\xspace}
\newcommand{\pascalfivei}{\mbox{PASCAL-$5^i$}}
\newcommand{\pascalteni}{\mbox{PASCAL-$10^i$}}
\newcommand{\coco}{\mbox{COCO-$20^i$}}
\newcommand{\newmetric}{Mean}
\newcommand{\trace}[1]{\text{Tr}(#1)}
\newcommand{\pold}{\pp^{\text{old}}}
\newcommand{\magicpar}[1]{\smallskip\noindent {\textbf{#1}}\enskip}

\newcommand{\appenref}[1]{\textcolor{red}{Appendix~\ref{#1}}}
\DeclareMathOperator*{\argmax}{argmax}

\begin{document}

\title{A Strong Baseline for Generalized Few-Shot Semantic Segmentation}
\author{
Sina Hajimiri\thanks{Corresponding author: seyed-mohammadsina.hajimiri.1@etsmtl.net}  \qquad
Malik Boudiaf \qquad Ismail Ben Ayed \qquad Jose Dolz \\
ÉTS Montreal 
}
\maketitle

\begin{abstract}
   This paper introduces a generalized few-shot segmentation framework with a straightforward training process and an easy-to-optimize inference phase. In particular, we propose a simple yet effective model based on the well-known InfoMax principle, where the Mutual Information (MI) between the learned feature representations and their corresponding predictions is maximized. In addition, the terms derived from our MI-based formulation are coupled with a knowledge distillation term to retain the knowledge on base classes.
   With a simple training process, 
   our inference model can be applied on top of any segmentation network trained on base classes.
   The proposed inference yields substantial improvements on the popular few-shot segmentation benchmarks, \pascalfivei{} and \coco{}. Particularly, for novel classes, the improvement gains range from 7\% to 26\% (\pascalfivei{}) and from 3\% to 12\% (\coco{}) in the 1-shot and 5-shot scenarios, respectively. Furthermore, we propose a more challenging setting, where performance gaps are further exacerbated. Our code is publicly available at \url{https://github.com/sinahmr/DIaM}.
\end{abstract}



\section{Introduction} \label{sec:introduction}

With the advent of deep learning methods, the automatic interpretation and semantic understanding of image content have drastically improved in recent years. These models are nowadays at the core of a broad span of visual recognition tasks and have enormous potential in strategic domains for our society, such as autonomous driving, healthcare, or security. Particularly, semantic segmentation, whose goal is to assign pixel-level categories, lies as one of the mainstays in visual interpretation. Nevertheless, the remarkable performance achieved by deep learning segmentation models is typically limited by the amount of available training data. Indeed, standard segmentation approaches are often trained on a fixed set of predefined semantic categories, commonly requiring hundreds of examples per class. This limits their scalability to novel classes, as obtaining annotations for new categories is a cumbersome and labor-intensive process.

Few-shot semantic segmentation (FSS) has recently emerged as an appealing alternative to overcome this limitation \cite{wang2019panet,tian2020prior,boudiaffew2021}. Under this learning paradigm, models are trained with an abundant labeled dataset on \textit{base} classes, and only a few instances of \textit{novel} classes are seen during the adaptation stage. However, \cite{tian2022generalized} identified two important limitations that hamper the application of these methods in real-life scenarios. First, existing literature on FSS assumes that the support samples contain the categories present in the query images, which may incur costly manual selection processes. Second, even though significant achievements have been made, all these methods focus on leveraging supports as much as possible to extract effective target information, but neglect to preserve the performance on known categories. Furthermore, while in many practical applications the number of novel classes is not limited, most FSS approaches are designed to work on a binary basis, which is suboptimal in the case of multiple novel categories.

Inspired by these limitations, a novel Generalized Few-Shot Semantic Segmentation (GFSS) setting has been recently introduced in \cite{tian2022generalized}. In particular, GFSS relaxes the strong assumption that the support and query categories are the same. This means that, under this new learning paradigm, providing support images that contain the same target categories as the query images is not required. Furthermore, the evaluation in this setting involves not only \textit{novel} classes but also \textit{base} categories, which provides a more realistic scenario. 

Although the setting in \cite{tian2022generalized} overcomes the limitations of few-shot semantic segmentation, we argue that a gap still remains between current experimental protocols and real-world applications. Hereafter, we highlight limiting points of the current literature and further discuss them in \cref{sec:practical_setting}.

\paragraph{Unrealistic prior knowledge.} We found that existing works explicitly rely on prior knowledge of the novel classes (supposed to be seen at test-time only) during the training phase. This, for instance, allows to filter out images containing novel objects \cite{tian2022generalized,lang2022learning} from the training set. Recent empirical evidence \cite{sun2022singular} found out that such assumptions indeed boost the results in a significant manner. 

\paragraph{Modularity.} Another limitation is the tight entanglement between the training and testing phases of current approaches, which often limits their ability to handle arbitrary tasks at test time. Specifically, existing meta-learning-based approaches are designed to handle binary segmentation \cite{lang2022learning}, and need to be consequently modified to handle multiple classes. While we technically address that by using multiple forward passes (one per class) followed by some heuristic aggregation of segmentation maps, this scales poorly and lacks principle. 

\paragraph{Contributions.} Motivated by these limitations, we aim to address a more practical setting and develop a fully modular inference procedure. Our inference abstracts away the training stage, making no assumption 
about the type of training or the format of tasks met at test time. Specifically: 

\begin{itemize}
    \item We present a new GFSS framework, \textit{\ourmethod{}} (Distilled Information Maximization). Our method is inspired by the well-known InfoMax principle, which maximizes the Mutual Information between the learned feature representations and their corresponding predictions. To reduce performance degradation on the base categories, without requiring explicit supervision, we introduce a Kullback-Leibler term that enforces consistency between the old and new model's base class predictions.
     
    \item Although disadvantaged by rectifications to improve the practicality of previous experimental protocols, we still demonstrate that \ourmethod outperforms current SOTA on existing GFSS benchmarks, particularly excelling in the segmentation of novel classes.
     
    \item Based on our observations, we go beyond standard benchmarks and present a more challenging scenario, where the number of base and novel classes is the same. In this setting, the gap between our method and the current GFSS SOTA widens, highlighting the poor ability of modern GFSS SOTA to handle numerous novel classes and the need for more modular/scalable methods.
\end{itemize}

\section{Related work}

\paragraph{Few-shot segmentation.} Few-shot semantic segmentation (FSS) has received notable attention in recent years, greatly inspired by the success of the few-shot learning paradigm \cite{finn2017model,ravi2016optimization}. Early FSS frameworks consisted in a dual-branch architecture, where one branch generated the class prototypes from support samples and the other one segmented the query images by exploiting the learned prototypes \cite{dong2018few,rakelly2018conditional,shaban2017oneshot}. Following the success of these pioneer approaches, an important body of literature explored how to better leverage category information from support samples to better guide the segmentation of query images \cite{li2021adaptive,tian2020prior,yang2020prototype,zhang2021self,liu2020part,wang2019panet}. For example, this can be achieved by collecting more abundant information from support images, which is used to construct multiple prototypes per class, each activating different regions of the query image \cite{li2021adaptive,yang2020prototype}. Alternative solutions to learn better category representations include: establishing correspondences between support and query images with Graph CNNs \cite{wang2020few}, imprinting the weights for novel classes \cite{siam2019amp}, or leveraging visual transformers to improve the category information transfer between support and query samples \cite{zhang2021few,liu2022intermediate,lu2021simpler}. Last, inspired by recent works in few-shot classification that favor a transductive setting, foregoing episodic training (\textit{aka} meta-learning) \cite{boudiaf2020information,dhillon2019baseline,liu2018learning,ziko2020laplacian}, RePRI \cite{boudiaffew2021} proposed a simple transductive solution.

\paragraph{Generalized few-shot segmentation.} To overcome some of the limitations of FSS, \cite{tian2022generalized} recently extended this setting, which was coined as generalized few-shot semantic segmentation (GFSS). 
In particular, GFSS approaches are given a single support set containing some images for every novel class, and they should be able to predict all potential base and novel classes in all query images. This way, in contrast to standard FSS methods, models have no knowledge of novel classes present in a query image. To tackle this problem, CAPL \cite{tian2022generalized} proposed a framework with two modules to dynamically adapt both base and novel prototypes. Nevertheless, the presented results are biased toward base classes and the solution requires that base classes are labeled in the support samples. 
Furthermore, the recent BAM model \cite{lang2022learning}, which was initially proposed for FSS, is also evaluated in the GFSS setting. This model consists of two steps. First, a \textit{base-learner} is trained on base classes following the standard supervised learning paradigm, where the cross entropy loss is employed on the base training set. Then, a second meta-learning step is introduced, where the base-learner and a new \textit{meta-learner} are optimized using episodic training. In the inference phase, the output of the meta-learner is fused with base-learner's output to give predictions on base classes and a single novel class. 
The fact that the meta-learner is only able to discern background-foreground categories makes this method's direct application not suitable to multi-class GFSS.

\section{Background}

\subsection{Preliminaries}
    \paragraph{Notations.} Let us note $H$ and $W$ the height and width of the original images, and $\Omega = [0, H-1] \times [0, W-1]$ the set of all pixels coordinates. In all generality, we define a segmentation model $f$ that takes images $\xx \in \mathbb{R}^{|\Omega| \times 3}$ as inputs, and produces segmentation maps $f(\xx)=\pp \in [0, 1]^{|\Omega| \times K}$, where $K$ denotes the number of classes to predict. 
    
    \magicpar{Standard few-shot segmentation.} In few-shot segmentation, two sets of classes are considered: the \textit{base} classes, $\Cb$, containing classes over which the model is trained; and the \textit{novel} classes, $\Cn$, strictly disjoint from base classes, such that \mbox{$\Cb \cap \Cn = \emptyset$}. At test time, the model is evaluated through a series of tasks. In each task, the model is given access to a support set $\support=\{\xx_i, \yy_i\}_{i=1}^{|\support|}$, containing a few images (shots), along with their corresponding binary segmentation masks $\yy_i \in [0, 1]^{|\Omega|}$, for some novel class randomly sampled from $\Cn$. Using this limited supervision, the model is then evaluated based on its ability to segment objects from this novel category in an unlabeled image, referred to as the \textit{query} image, which will be referred to as the $(|\support|+1)^{th}$ image. In this context, $K=2$ and the model is expected to produce a binary mask $\pp \in [0, 1]^{|\Omega| \times 2}$.
    
    \magicpar{Generalized few-shot segmentation.} The \textit{generalized} setting extends the standard setting to account for the fact that real-world applications often require being able to recognize both base and novel classes in new images. 
    As a result, for each task, we now require the model to produce a segmentation over the $1+\Nb+\Nn$ (including the background) potential classes, such that
    $\pp \in [0,1]^{|\Omega|\times (1+\Nb+\Nn)}$. So, for a given pixel $j$ we have
    \begin{align}
        \phantom{,}
        \pp(j) = \left[ \overbrace{p_0}^{\text{bg}}, ~
        \overbrace{p_1, \dots, p_{\Nb}}^{\text{base classes}},~ \overbrace{p_{\Nb+1}, \dots, p_{\Nb+\Nn}}^{\text{novel classes}} \right]^\top,
    \end{align}
    in which we omit pixel index $j$ from the right-hand side for simplicity, and \textit{bg} stands for the background.

\subsection{Toward a fully practical setting} \label{sec:practical_setting}
    As motivated in \cref{sec:introduction}, we aim at evaluating methods in a maximally practical setting. Therefore, we start by rectifying design choices made in previous works \cite{tian2022generalized,lang2022learning} that we find impractical and that could impact the results in a significant manner. We further posit additional desiderata to improve the practicality of developed methods and widen their adoption.
    
    \paragraph{Addressing the presence of novel classes during training.} We found that previous works \cite{tian2022generalized,lang2022learning} explicitly removed images containing novel classes during the training phase. This implicitly requires information that should not be available at that stage, namely the prior knowledge of the novel classes, as well as the potential presence of a given novel object in a particular image. \emph{Instead, in our setting, we keep those images during training (as they naturally occur), and the potential objects from novel classes are labeled as background at that stage.} Needless to say, this may negatively impact the performance of the model at test time, given the class ambiguity introduced while forcing the network to predict potential novel classes as background. However, we believe that this is a more natural way to design the problem.

    \paragraph{Relaxing test-time labeling requirements.} We found that previous works \cite{tian2022generalized} required potential objects from base classes to be explicitly labeled in the images of the support set $\support$. We argue that this can require a significant additional load of work in real-world settings. Consider the simple example of \coco{}'s dataset, with a total of $80$ classes. Instead of only annotating objects from $20$ novel classes, as was the case in the standard FSS setting, annotators would now have to search for $60$ potential additional classes in each image from the support, thus leading to a substantial increase in human/financial efforts. We find that this requirement is not necessary, and high performances on base classes can be retained by other means. \emph{Therefore, we drop this requirement, and only require annotations for novel classes at test-time, whereas the rest (including potential objects from base classes) are labeled as background}.

    \paragraph{Modularity of inference.} Although not a strict requirement, we advocate developing modular inferences that, unlike current approaches \cite{lang2022learning,tian2022generalized}, can apply to any model, without relying on customized architectures or training procedures. The rationale is two-fold. First, as foundation models are and will continue to push state-of-the-art on most vision tasks, we forecast that the ability to leverage off-the-shelf models seamlessly will become crucial in reaching high performances. Second, it drastically lowers the entry barrier to few-shot learning for practitioners who, in most cases, already possess trained models for their specific application, and whose limited computational resources may prevent re-training models for every method they would like to try. \emph{Access to inference-only methods which can readily equip preexisting models with few-shot ability could be a key ingredient for the widespread adoption of few-shot methods}.

\section{Our method}

In light of the requirements and desiderata posed in \cref{sec:practical_setting}, we shift our attention from training to inference. In particular, unlike previous generalized FSS methods, we use a standard supervised training procedure that is not informed by any knowledge, even implicit, of the novel classes. We further develop an optimization-based inference procedure that can be directly deployed at test time. Each of these steps is detailed below.

\subsection{Training}
    For convenience, we partition the segmentation model into a feature extractor $f_\phi$ and a linear classifier $f_{\thetabase}$, trained in a standard supervised fashion to segment base classes $\Cb$ during training. At this stage, the classifier can only predict $1 + \Nb$ classes, \ie, the background and the base classes.
    
\subsection{Inference}
    At test-time, given $\Nn$ novel classes to recognize, we freeze the feature extractor $f_{\phi}$ and augment the pre-trained classifier $\thetabase \in \mathbb{R}^{(1+\Nb) \times d}$ with novel prototypes $\thetanovel \in \mathbb{R}^{\Nn \times d}$. We consider the concatenation $\thetabf=[\thetabase; \thetanovel] \in \mathbb{R}^{(1+\Nb+\Nn) \times d}$ to form our final classifier, and optimize $\thetabf$ for this specific task. Note that $d$ is the size of the feature space. We base our optimization objective on the seminal concept of mutual information \cite{shannon1948mathematical}. Specifically, we use the InfoMax framework \cite{linsker1988self} as a starting point for our formulation. InfoMax advocates maximizing the mutual information between a network's inputs and outputs as
    \begin{align} \label{eq:infomax}
        \phantom{,}
        \max_{\thetabf} ~ \mi{X}{P} = \underbrace{\ent{P}}_{\text{marginal entropy}} - \underbrace{\ent{P|X}}_{\text{conditional entropy}},
    \end{align}
    where $X$ and $P$ are the random variables respectively associated with the pixel distribution and model's predictions. Instantiated in our context, InfoMax \eqref{eq:infomax} incites the model to produce confident predictions on each pixel (conditional entropy), while encouraging an overall balanced marginal distribution (marginal entropy), \ie, roughly speaking, an equal number of pixels assigned to each class. Interestingly, in the related context of classification, InfoMax can be interpreted as an unsupervised clustering criterion \cite{jabi2019deep}.

    In the following sections, we explain how we gradually depart from the vanilla InfoMax principle and incorporate problem-specific constraints and inductive biases to reach our final formulation.

    \subsubsection{Enforcing high-confidence} \label{sec:high_confidence}
        We start by focusing our attention on the conditional entropy term mentioned in \cref{eq:infomax} that enforces
        high-confidence predictions for each pixel. To this end, we introduce the cross-entropic operator:
        \begin{align}
            \phantom{.}
            \ent{\pp_i;\qq_i} = \frac{-1}{|\Omega|} ~\trace{\pp_{i} \log(\qq_{i}^{\top})}.
        \end{align}
        Taking into account support and query images, the vanilla conditional entropy of $P|X$ can be written as
        \begin{align} \label{eq:vanilla_ent}
            \phantom{.}
            \ent{P|X} = \frac{1}{|\support|+1} \sum_{i=1}^{|\support|+1} \ent{\pp_i; \pp_i}.
        \end{align}
        
        \paragraph{Leveraging supervision} Entropy $\ent{\pp}$ can be interpreted as a self-cross-entropy $\ent{\pp; \pp}$, in which the model's own predictions are used as pseudo-labels for supervision. Because actual ground-truth labels for the support images are provided, we can effectively replace those pseudo-labels with the ground truth.
        
        \paragraph{Aligning support labels and predictions.} As motivated in \cref{sec:practical_setting}, we do not require objects from base classes to be labeled in the support images. That produces a slight misalignment between labels and the model's predictions. More specifically, a pixel $j$ labeled as \textit{background}, $\yy_{i}(j)=[1, 0, \dots, 0]$, can now have two meanings: either it actually is a
        background pixel or it belongs to a base class object. To account for that misalignment between predictions $\pp$ and labels $\yy$, we project the model's predictions as
        \begin{align} \label{eq:p-ce}
            \pi_{\support}(\pp_i)(j) = \left[ \sum_{k=0}^{\Nb} p_k, \; \overbrace{0, \dots, 0}^{\text{$\Nb$ times}}, \; p_{\Nb+1}, \dots, p_{\Nb + \Nn} \vphantom{\sum_{i=1}^{\Nb}\overbrace{\dots}^{\text{$\Nb$}}}\right]^\top.
        \end{align}
        We can now write our constrained conditional entropy, partitioning pixels into supervised and unsupervised. We found it beneficial to adjust the relative weighting of the terms, and therefore introduce $\alpha>0$ in the objective, such that our conditional entropy term reads as
        \begin{align} \label{eq:constrained_ent}
            \mathcal{L}_{\text{cond-ent}} = \alpha \underbrace{ \sum_{i=1}^{|\support|} \ent{\yy_i; \pi_{\support}(\pp_i)}}_{\mathcal{L}_\text{xent}: \text{ Support supervised entropy}}  + \underbrace{ ~ \ent{\pp_{|\support|+1}},}_{\text{Query unsupervised entropy}}
        \end{align}
        where $\alpha$ controls the reliance on the labeled support set. We will refer to the \textit{support supervised entropy} term as $\mathcal{L}_\text{xent}$.
        
    \subsubsection{Addressing class imbalance}
        The constraint of high-confidence predictions alone can be easily satisfied by a model and does not constrain the problem enough to guarantee meaningful solutions. For instance, a trivial classifier assigning all pixels from the query image to the same class with maximum probability 1 would fully satisfy the high-confidence constraint. Therefore, we need to go beyond mere entropy minimization, which naturally leads us to shift our attention to the marginal entropy term from \cref{eq:infomax}. As shown below, marginal entropy ensures a
        fair distribution of assignments over the different classes, thereby preventing, \eg, the trivial solutions previously described. As in the previous section, let us start with the vanilla formulation of the marginal entropy
        \begin{align} \label{eq:marginal_and_kl}
            \phantom{,}
            \ent{P} = - \widehat{\pp} \cdot \log(\widehat{\pp}) =  \text{Cste} - \kl{\widehat{\pp}}{\mathbf{u}},
        \end{align}
        where $\cdot$ is the dot-product, $\kl{.}{.}$ denotes the Kullback-Leibler divergence, $\mathbf{u}= 1 / (1+\Nb + \Nn ) \cdot \mathbf{1}$ is the uniform distribution, and $\widehat{\pp} \in [0, 1]^{1+\Nb + \Nn}$ is the model's marginal distribution over classes (detailed hereafter). In other words, $\ent{P}$ regularizes the overall procedure by encouraging class-balanced predictions, \ie, predicting roughly an even distribution of pixels for each class. In the context of our problem, this vanilla formulation exhibits important limitations. \\

        First, akin to the previous section, we have direct access to
        supervision for the support pixels. Assuming that the model fits those labels well, the distribution of predictions will naturally converge to the proportions dictated by the ground truth labels. Therefore, we do not include the predictions
        from the support samples when computing the marginal, as this would provide a redundant signal, and only consider the marginal distribution over the query:
        \begin{align}
            \phantom{.}
            \widehat{\pp} = \frac{1}{|\Omega|} \sum_{j \in \Omega} \pp_{|\support|+1}(j).
        \end{align}

        Additionally, semantic segmentation is virtually never a balanced problem. The number of instances, the distance to the camera, or the angle of view are all factors that can randomly vary between scenes, significantly affecting the final share of pixels that each class occupies in a given frame. Therefore, using the uniform distribution as a prior to match can be sub-optimal, as shown in our ablation study in \cref{sec:ablation_study}. Beyond merely down-weighting to weaken this regularization, alternatives in the literature include using $\alpha$-entropy \cite{veilleux2021realistic} in place of the standard Shannon entropy from \cref{eq:marginal_and_kl}, or replacing $\textbf{u}$ with an estimated prior $\prior$ \cite{boudiaffew2021}. We decided to go with the prior estimation procedure given in \cite{boudiaffew2021}. Specifically, we extend the marginal entropy to take into account a prior:
        \begin{align} \label{eq:l-marg-ent}
            \phantom{.}
            \mathcal{L}_{\text{marg-ent}} = \ent{P; \prior} = \text{Cste} - \kl{\widehat{\pp}}{\prior}.
        \end{align}
        This \textit{prior-guided} marginal entropy loss reduces to the standard marginal entropy in the absence of prior, \ie, $\prior=\textbf{u}$. Following \cite{boudiaffew2021}, $\prior$ is estimated from the model's initial marginal distribution and re-updated during optimization.

    \subsubsection{Preserving base knowledge} \label{sec:kd}
         So far, our inference procedure has not made any distinction between base and novel classes, thus leaving aside an important inductive bias of our problem: prototypes from the base classifier, $\thetabase$, were trained using orders of magnitude more data than prototypes from novel classes, whose only available supervision comes from the few labeled samples from the support set. Additionally, our setting prevents support images from providing any explicit supervision for base classes, which only accentuates the asymmetry between base and novel classes. \\
         
         To account for these two contrasts, a simple solution could be to freeze the base classifier $\thetabase$, and only optimize the novel classifier $\thetanovel$. However, we show in \cref{sec:experiments} that this results in sub-optimal results. Instead, we propose a more flexible self-distillation term that encourages the model's predictions on base classes to stay close to its old predictions. Formally, we consider the base classifier's weights $\thetabase^{(0)}$ (right after training) and define the model's \textit{old} predictions as
         \begin{align}
            \phantom{.}
             \pold_{i} = f_{\thetabase^{(0)}} \circ f_{\phi}(\xx_i) \in [0, 1]^{|\Omega| \times (1+\Nb)}.
         \end{align}

         \paragraph{New-to-old mapping.} In order to measure and minimize any sort of distances between the old model's predictions $\pold_i$ defined over base classes and our current model's predictions $\pp_i \in [0, 1]^{|\Omega| \times (1+\Nb+\Nn)}$, defined over both base and novel classes, we must map them to the same label space. At this point, it is important to recall that a \textit{background} prediction from the base model really means ``anything other than base classes". That includes actual background, as well as potential novel classes that were labeled as background during training. 
         Therefore, to make $\pp$ and $\pold$ consistent, we project $\pp$ as
         \begin{align}
             \pi_{\text{new2old}}(\pp)(j) = {\left[
        	p_0 + \sum_{i = 1}^{\Nn} p_{\Nb+i}, \;
        	p_1, p_2, \dots, p_\Nb \right]}^\top.
         \end{align}
         
         Now, inspired by recent literature in incremental learning that distills knowledge using the predictions of old models \cite{castro2018end,dong2022federated,li2017learning,cermelli2020modeling}, we
         can express our \textit{knowledge-distillation} term applied to the query image as
         \begin{align} \label{eq:kd}
            \phantom{,}
             \mathcal{L}_{\text{KD}} = \kl{\pi_{\text{new2old}}(\pp_{|\support|+1})}{\pold_{|\support|+1}} ,
         \end{align}
         which enables us to write our final objective:
         \begin{align} \label{eq:final}
            \phantom{.}
            \min_{\thetabf} \quad \mathcal{L}_{\ourmethod} = \mathcal{L}_{\text{cond-ent}} - \mathcal{L}_{\text{marg-ent}} + \beta \mathcal{L}_{\text{KD}},
         \end{align}
         where $\beta$ controls the importance of retaining base classes knowledge.

\section{Experiments} \label{sec:experiments}

\subsection{Experimental setting} \label{sec:exp-setting}
    \paragraph{Datasets.} To evaluate our method we use two well-known few-shot segmentation benchmarks:
    \pascalfivei{} \cite{everingham2015pascal,hariharan2014simultaneous,shaban2017oneshot} and \coco{} \cite{lin2014microsoft,shaban2017oneshot}.
    To use in our experiments in \cref{table:balanced-pascal}, we define \pascalteni{} in the same way \pascalfivei{} is formed \cite{shaban2017oneshot}, but splitting the set of classes into two subsets of size 10, instead of four subsets of size 5. More specifically, in \pascalteni{}, the subset $i$ consists of classes with indices \mbox{$\{10i + j\}$} for \mbox{$j \in \{1, 2, \dots, 10\}$}. 
    For \coco{} we report the average performance of models over 10K query images, while for \pascalfivei{} and \pascalteni{} we use all available query images. More specifications about the datasets can be found in \appenref{app:datasets}. 

    \paragraph{Evaluation protocol.} For evaluation purposes, we resort to the standard mean intersection-over-union (mIoU) over the classes. In our tables, \textit{Base} and \textit{Novel} refer to mIoU over base and novel classes, respectively. 
    Although mIoU over all classes has been used by prior works \cite{tian2022generalized,lang2022learning} as the overall score, we believe it is a misleading metric in GFSS. Classes in \pascalfivei{} and \coco{} are split in a way that the number of base classes is thrice the number of novel classes. This biases the metric toward the \textit{Base} score and undermines the very goal of few-shot learning, which is learning the novel classes. Therefore, we propose to use the standard average of \textit{Base} and \textit{Novel} scores as the \textit{\newmetric} score in the GFSS task.
    Following \cite{tian2022generalized, lang2022learning}, metrics are averaged over 5 independent runs. 
    
    \paragraph{Implementation details.}
    The architecture of the model is based on PSPNet \cite{zhao2017pyramid} using Resnet-50 \cite{he2016deep} backbone. 
    During training, a standard cross-entropy over the base classes is minimized.
    Our training scheme follows the base-training stage of \cite{lang2022learning}, so, the batch size is 12 and SGD optimizer is used with an initial learning rate $2.5 \times 10^{-4}$, momentum $0.9$, and weight decay $10^{-4}$. The number of epochs is 20 for \coco{} and 100 for \pascalfivei{} and \pascalteni{}. Data augmentation is done in the same way as \cite{tian2020prior}.
    At inference time, the feature extractor $\phi$ is kept frozen and the classifier $\thetabf$ is optimized. For this phase, SGD optimizer is used with learning rate $1.25 \times 10^{-3}$ and the loss function in \cref{eq:final} is optimized for 100 iterations. 
    The size of the feature space $d$ is set to 512 in all experiments. We have empirically found that $\mathcal{L}_\text{xent}$ in \cref{eq:constrained_ent} and $\mathcal{L}_{\text{KD}}$ of \cref{eq:kd} play more significant roles in the model's performance, and upweighted these terms by two orders of magnitude ($\alpha = \beta = 100$).
    Following \cite{boudiaffew2021}, the value of $\prior$ is estimated by the model at the beginning of the evaluation and it is updated once at iteration 10. 
    
    \paragraph{Baselines.} Following \cite{tian2022generalized}, we included relevant FSS methods in our evaluation, including CANet \cite{zhang2019canet}, PANet \cite{wang2019panet}, PFENet \cite{tian2020prior}, and SCL \cite{zhang2021self}. These methods were adapted in \cite{tian2022generalized} by modifying their respective inference code to generate prototypes for both base and novel classes in each query image. 
    We also modified the inference code of RePRI \cite{boudiaffew2021} to accommodate multiple classes during testing and adapted MiB \cite{cermelli2020modeling}, an incremental learning method, to the GFSS setting, as a distillation term similar to \cref{eq:kd} is integrated in their approach.
    Furthermore, we compare our method to the GFSS method CAPL \cite{tian2022generalized}.
    Last, although BAM \cite{lang2022learning} reports results in the GFSS task, its episodic learning nature hinders the scalability of this approach to settings where segmentation of multiple novel classes is required. Indeed, at inference, it can only provide background-foreground predictions, which is impractical in our current validation. In order to include it in our experiments, we have made some changes to this method, which are detailed in \appenref{app:bam-adaptation}. Note that the reported results in \cref{table:all-main} do not take into account the background IoU in the evaluation metrics, as this class is not an object of interest. This is discussed in more detail in \appenref{app:including-bg}.

    \begin{table*}[tbh]
    \centering
    \footnotesize
    \begin{tabular}{@{}l@{\hskip 3px}lccccccccc@{}}
        \toprule
        &&&& \multicolumn{7}{c}{\textbf{\pascalfivei{}}}\\
        &&&& \multicolumn{3}{c}{1-Shot} && \multicolumn{3}{c}{5-Shot} \\
        \cmidrule{5-11}
        Method && Practical setting & Multi-class design  & Base & Novel & \newmetric{} && Base & Novel & \newmetric{} \\
        \midrule
        CANet\textsuperscript{*} \cite{zhang2019canet} & {\scriptsize CVPR'19} & \cmark & \xmark & 8.73 & 2.42 & 5.58    &&    9.05 & 1.52 & 5.29 \\
        PANet\textsuperscript{*} \cite{wang2019panet} & {\scriptsize ICCV'19} & \cmark & \xmark & 31.88 & 11.25 & 21.57     &&    32.95 & 15.25 & 24.1\\
        PFENet\textsuperscript{*} \cite{tian2020prior} & {\scriptsize TPAMI'20} & \cmark & \xmark & 8.32 & 2.67 & 5.50     &&  8.83 & 1.89 & 5.36\\
        MiB\textsuperscript{\textdagger} \cite{cermelli2020modeling} & {\scriptsize CVPR'20} & \cmark & \cmark & 63.80 &  8.86 & 36.33 && 68.60 & 28.93 & 48.77 \\
        SCL\textsuperscript{*} \cite{zhang2021self} & {\scriptsize CVPR'21} & \cmark & \xmark & 8.88 & 2.44 & 5.66    &&    9.11 & 1.83 & 5.47 \\
        RePRI\textsuperscript{\textdagger} \cite{boudiaffew2021} & {\scriptsize CVPR'21} & \cmark & \xmark & 20.76 & 10.50 & 15.63    &&    34.06 & 20.98 & 27.52 \\
        \midrule
        CAPL\textsuperscript{\textdagger} \cite{tian2022generalized}  & {\scriptsize CVPR'22} & \xmark & \cmark & 64.80 & 17.46 & 41.13  &&    65.43 & 24.43 & 44.93\\
        BAM\textsuperscript{\textdagger} \cite{lang2022learning}  & {\scriptsize CVPR'22} & \xmark & \xmark & \textbf{71.60} & 27.49 & 49.55     &&    \textbf{71.60} & 28.96 & 50.28 \\
        \ourmethod{} & {\scriptsize (Ours)} & \cmark & \cmark & 70.89 & \textbf{35.11} & \textbf{53.00} && 70.85 & \textbf{55.31} & \textbf{63.08} \\
        \midrule
        \ourmethod-UB & {\scriptsize (Ours)} & \xmark & \cmark & 71.13 & 52.61 & 61.87 && 71.12 & 66.12 & 68.62\\
        \midrule
        &&&  & \\
        &&&& \multicolumn{7}{c}{\textbf{\coco{}}} \\
        \cmidrule{5-11}
        &&&& Base & Novel & \newmetric{} && Base & Novel & \newmetric{} \\
        \midrule
        RePRI\textsuperscript{\textdagger} \cite{boudiaffew2021} & {\scriptsize CVPR'21} & \cmark & \xmark & 5.62 & 4.74 & 5.18 &&   8.85 & 8.84 & 8.85 \\
        CAPL\textsuperscript{\textdagger} \cite{tian2022generalized} & {\scriptsize CVPR'22} & \xmark & \cmark & 43.21 & 7.21 & 25.21 &&   43.71 & 11.00 & 27.36 \\
        BAM\textsuperscript{\textdagger} \cite{lang2022learning} & {\scriptsize CVPR'22} & \xmark & \xmark & \textbf{49.84} & 14.16 & 32.00 && \textbf{49.85} & 16.63 & 33.24\\
        \ourmethod{} & {\scriptsize (Ours)} & \cmark & \cmark & 48.28 & \textbf{17.22} & \textbf{32.75} && 48.37 & \textbf{28.73} & \textbf{38.55} \\
        \midrule
        \ourmethod-UB & {\scriptsize (Ours)} &  \xmark & \cmark & 48.55 & 29.48 & 39.02 && 48.63 & 40.43 & 44.53 \\
        \bottomrule
    \end{tabular}

    \caption{\textbf{Quantitative evaluation on \pascalfivei{} and \coco{} compared to FSS and GFSS methods.} \ourmethod{} represents our method and \ourmethod-UB is an impractical extension of it, explained in \cref{sec:ablation_study}. All the methods employ ResNet-50 as backbone. Results with a ``*" sign are obtained from \cite{tian2022generalized}, whereas results with a ``\textdagger" sign are reproduced using the publicly available codes.}
    \label{table:all-main}
    \end{table*}

\subsection{Main results}
    \paragraph{Comparison to state-of-the-art GFSS.}
    \Cref{table:all-main} reports the results obtained by different approaches in the GFSS setting. Here, we stress some of the underlying limitations present in these methods. First, we refer to as `Practical setting' the scenario where models employ the whole dataset during the training of base classes, \ie, have no access beforehand to information about novel categories (more information is available in \appenref{app:practical}). Then, we resort to `Multi-class design' to highlight which methods, by nature, can handle multiple novel classes simultaneously.
    From these results, we can observe that models designed specifically for the task of FSS are unable to perform satisfactorily in both base and novel categories. 
    Compared to GFSS methods, our formulation brings substantial improvements under both the 1-shot and 5-shot scenarios, particularly on \textit{Novel} metric. More specifically, compared to CAPL, these differences are considerably large in the case of novel classes, with around 17\% and 31\% improvement on \pascalfivei. We believe that this imbalanced behavior might be due to the fact that CAPL relies on the base classes to generate the prototypes for novel categories. Thus, the model may give excessive importance to base classes, not fully leveraging the support samples during the adaptation stage.
    Furthermore, differences with respect to BAM are also significant, with 7\% and 26\% improvement on novel classes in the 1-shot and 5-shot settings on \pascalfivei.
    Note that BAM fuses the output on base classes to the novel prediction and it does not form a holistic classifier over all the classes \cite{lang2022learning}. Therefore, its prediction on base classes remains intact after learning the novel class, and this leads to high performance on \textit{Base} metric. However, we believe a rigid base prediction hampers the ability of the model to grasp a universal view of the classes, and this comes at the price of worse performance on novel classes, despite the fact that learning them is the main objective of any few-shot learning framework.
    
    The same trend observed in the \pascalfivei{} benchmark is repeated for \coco{}. More concretely, our method achieves performance gains of around 10\% and 17\% on \textit{Novel} metric over CAPL, and 3\% and 12\% compared to BAM under the 1-shot and 5-shot scenarios, respectively.
    
    \paragraph{Impact of increasing the number of novel classes.} 
    A more challenging scenario involves expanding the set of novel classes. To this end, we compare our approach to CAPL \cite{tian2022generalized} and BAM \cite{lang2022learning} on the newly defined \pascalteni{}, which contains 10 base and 10 novel classes. \Cref{table:balanced-pascal} shows that the difference in \textit{Novel} scores compared to these methods is further exacerbated. In particular, \ourmethod{} yields improvements on novel classes of nearly 15\% and 30\% under 1-shot and 5-shot settings, respectively, while also outperforming both methods on base classes.

    \begin{table}[ht!]
    \centering
    \resizebox{1.0\linewidth}{!}
    {
    \begin{tabular}{@{}l@{\hskip 3px}lcccccccc@{}}
        \toprule
        &&& \multicolumn{3}{c}{1-Shot} && \multicolumn{3}{c}{5-Shot} \\
        \cmidrule{4-10}
        Method &&& Base & Novel & \newmetric && Base & Novel & \newmetric \\
        \midrule
        CAPL \cite{tian2022generalized} & {\scriptsize CVPR'22} && 53.78 & 15.01 & 34.40 && 57.02 & 20.40 & 38.71 \\
        BAM \cite{lang2022learning} & {\scriptsize CVPR'22} && 69.02 & 15.48 & 42.25 && 69.18 & 21.51 & 45.35 \\
        \ourmethod{} & {\scriptsize (Ours)} && \textbf{70.26} & \textbf{31.29} & \textbf{50.77} && \textbf{70.25} & \textbf{51.89} & \textbf{61.07} \\
        \bottomrule
    \end{tabular}
    }
    \caption{\textbf{Quantitative evaluation on \pascalteni{}.} All the methods employ ResNet-50 as backbone.}
    \label{table:balanced-pascal}
    \end{table}

\subsection{Ablation studies} \label{sec:ablation_study}
\begin{figure}[t]
        \centering
        \includegraphics[width=\columnwidth]{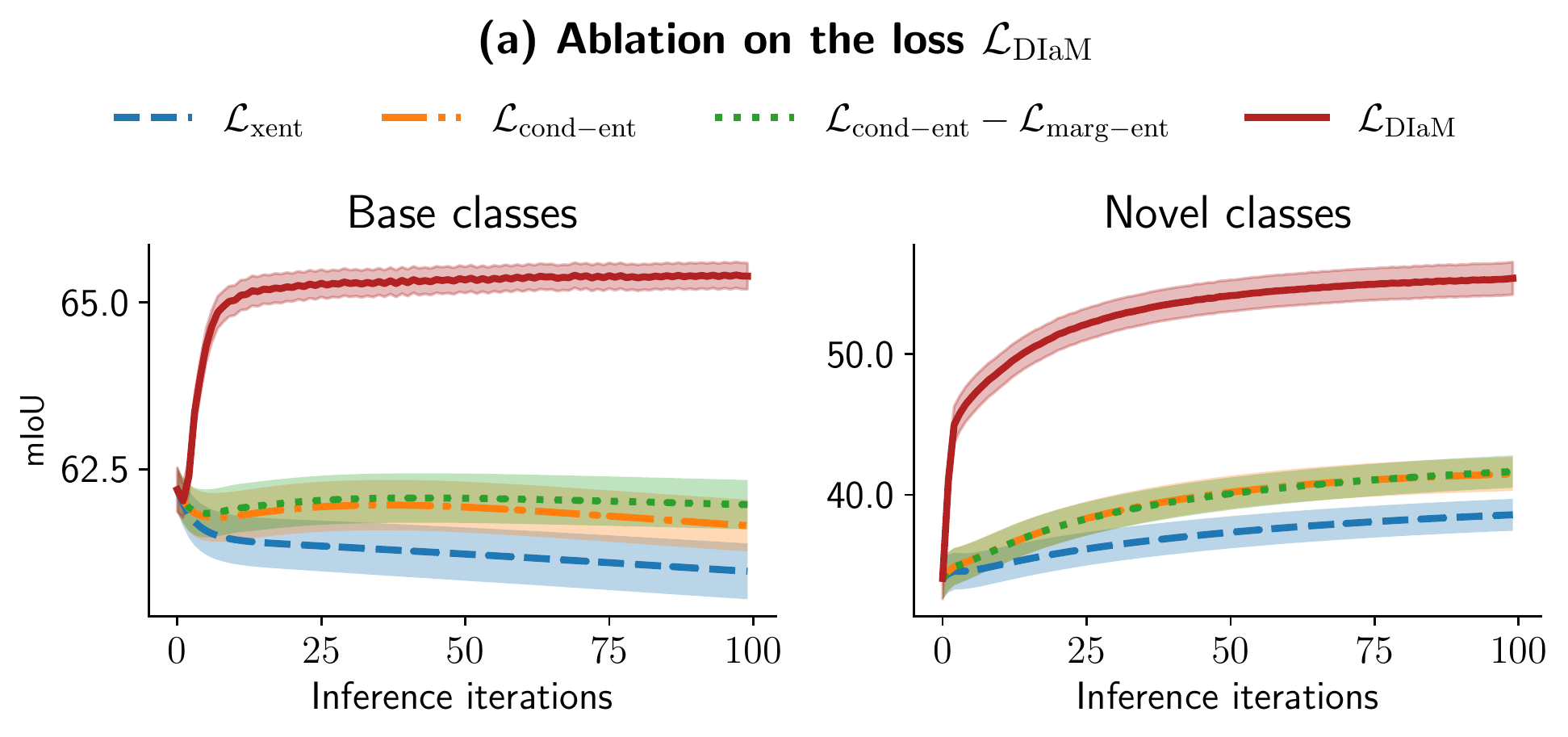}
        \includegraphics[width=\columnwidth]{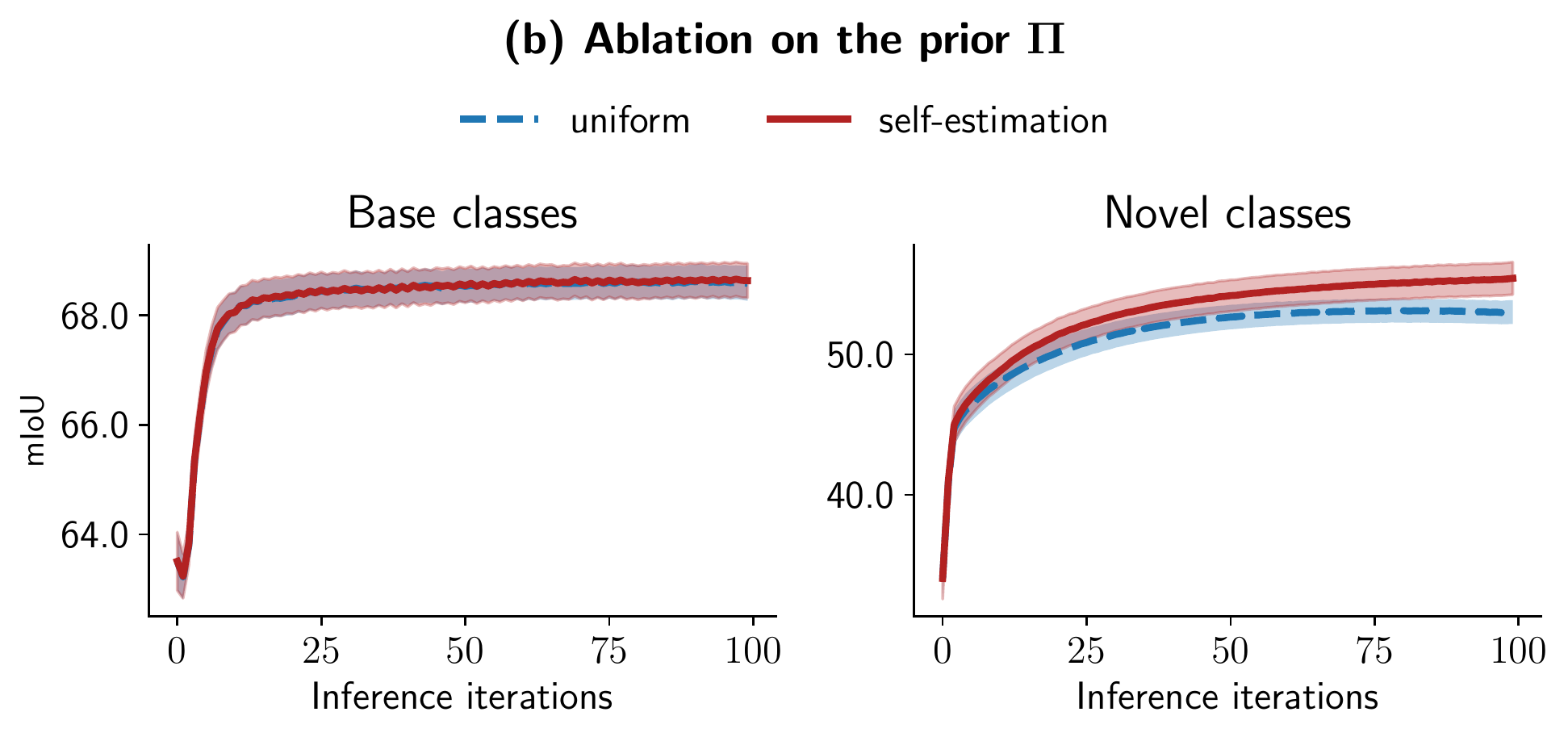}
        \includegraphics[width=\columnwidth]{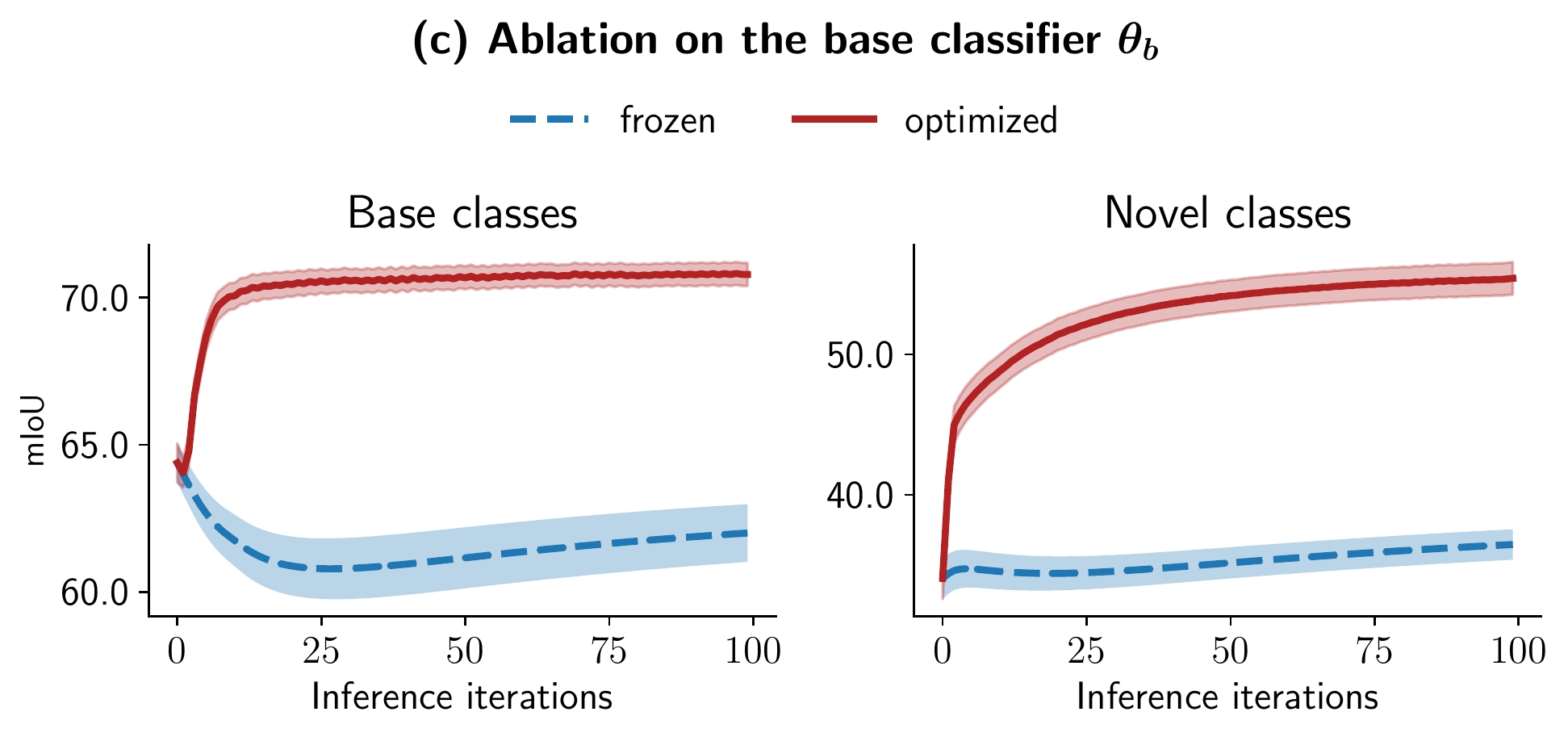}
        \caption{\textbf{Ablation studies} on (a) \ourmethod's loss, (b) the type of prior $\prior$ in \cref{eq:l-marg-ent}, and (c) the optimization of $\thetabase$. Results are provided for \pascalfivei{} under the 5-shot setting.}
        \label{fig:ablation_study}
    \end{figure}

    We ablate along three axes to better understand each design choice's contribution. Results are summarized in the form of convergence plots in \cref{fig:ablation_study}. Specifically, we find that (a) terms act symbiotically to provide the best performance on both base and novel classes, (b) self-estimation yields higher novel-class performance than the uniform prior, and finally, (c) as stated in \cref{sec:kd}, introducing the knowledge distillation term and optimizing the base classifier $\thetabase$, as opposed to the simple solution of freezing it, allows both faster convergence and better optima. 
    Note that since we form a holistic distribution over all classes, even if we freeze $\thetabase$, the probability of base classes will not remain fixed. Those convergence plots demonstrate that models improve rapidly at first and keep improving at a slower pace as the adaptation continues (more details in \appenref{app:num-iter}).

    Based on \cref{fig:ablation_study}, the self-estimation of $\prior$ leads to better performance than using the uniform distribution. Therefore, we wonder: \textit{how far could this term take us?} The methods mentioned as \ourmethod-UB in \cref{table:all-main} demonstrate an extension of our model in which the actual true $\prior$ is given. This shows the upper bound of the performance of our model as the estimation of $\prior$ gets more accurate. Experiments show that an accurate $\prior$ can lead to significant improvements in performance.
    We emphasize that knowing the exact size of the target objects might be unrealistic and that our goal is just to demonstrate that with a proper mechanism to provide an accurate estimate of the target class proportion, the obtained results can be much improved.

\subsection{Visual examples} \label{sec:visual}
\begin{figure}[tb]
        \centering
        \begin{subfigure}[t]{\linewidth}
            \centering
            \includegraphics[width=\linewidth]{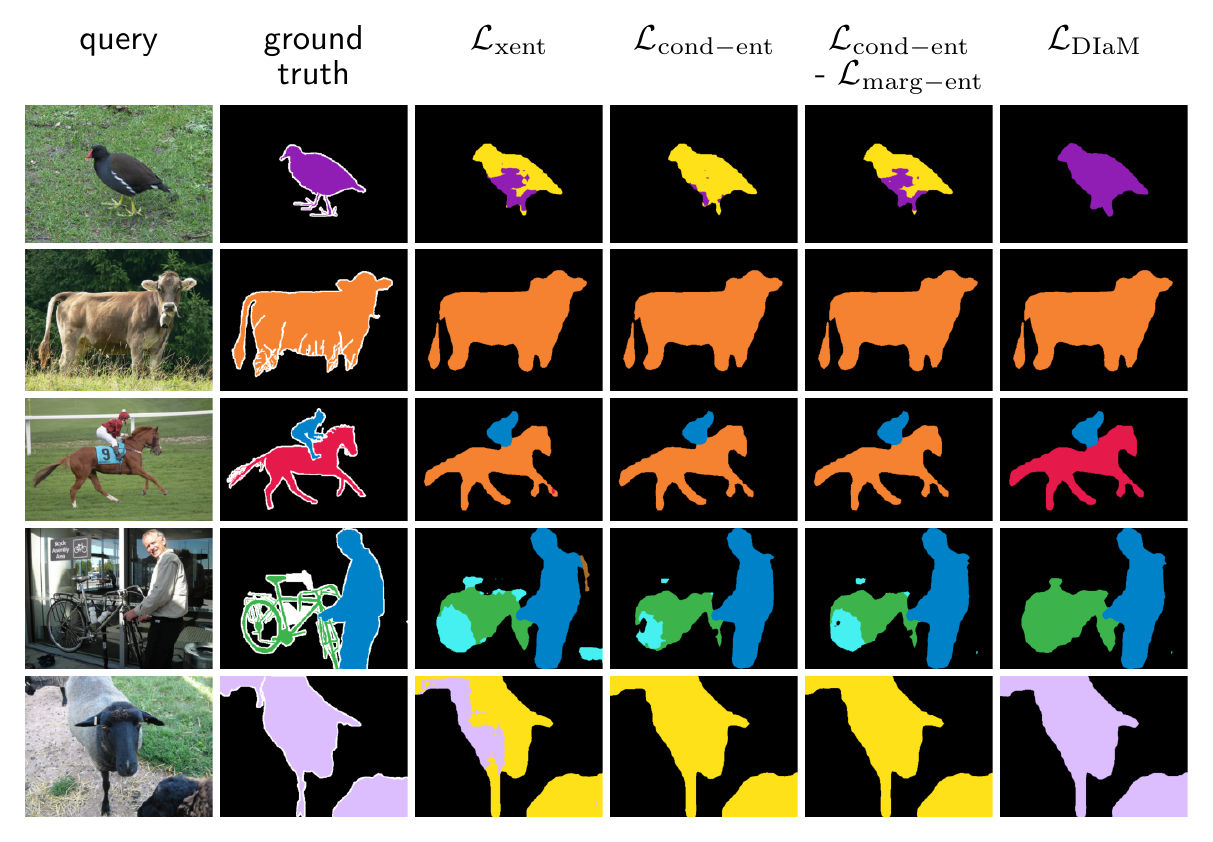}
        \end{subfigure}
        \begin{subfigure}[t]{\linewidth}
            \centering
            \includegraphics[width=0.9\linewidth]{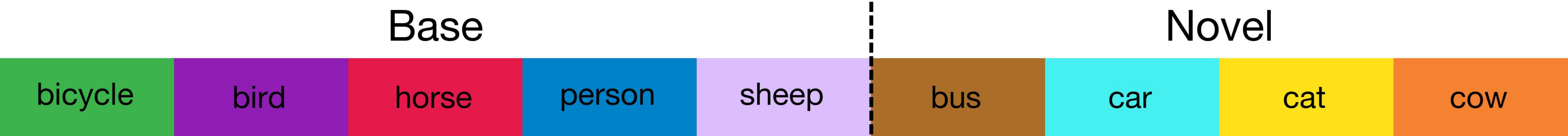}
        \end{subfigure}
        \caption{\textbf{Qualitative results of different terms of \ourmethod's loss function (on \pascalfivei{} under the 5-shot setting).} A single support set, containing novel classes \textit{bus}, \textit{cat}, \textit{car}, \textit{chair}, and \textit{cow}, is used for predicting every query image. Query images can contain any classes and every one of them is to be recognized.}
        \label{fig:visual}
    \end{figure}

    Qualitative results of the proposed method are presented in \cref{fig:visual}. More specifically, for a given query, the predictions made by four different models, each containing a subset of our loss function are shown. This figure reveals an interesting phenomenon: in the absence of the knowledge distillation term of \cref{eq:kd}, the model tends to predict some of the base classes as the novel ones. For example, in the third and fourth row, the base classes \textit{horse} and \textit{bicycle} are mistakenly segmented as the novel classes \textit{cow} and \textit{car}. These misclassifications are revised when the knowledge distillation term is in use, which further proves its effectiveness. More visual examples are provided in \appenref{app:visual}.

\section{Conclusion}

\paragraph{Summary.} We propose a new generalized few-shot segmentation method, with a standard supervised training scheme and a lightweight inference phase, which can be applied on top of any learned feature extractor and classifier. Our method is based on the InfoMax framework \cite{linsker1988self} incorporating problem-specific biases, and it also employs knowledge distillation \cite{hinton2015distilling} to prevent performance loss on the classes learned during training. Compared to prior works, our results show significant improvement in learning novel classes, while keeping the performance on base classes high as well. 
We eliminated some limiting assumptions of prior methods, such as recognizing one novel class at a time, benefiting from some information about novel classes during training, and having to label base classes in support images. Our proposed knowledge distillation considerably helps retain base knowledge, and we believe imposing such a term is more realistic and practical than explicit supervision for base classes.

\paragraph{Limitations.} Results of the \ourmethod-UP experiments show that our marginal entropy term in \cref{eq:l-marg-ent} can play a significant role and increase the performance considerably. In particular, the ablation study demonstrates that even though our simple choice of estimating the prior proportion $\prior$ using the model's predictions introduces slight improvements, access to a more precise prior has the potential to substantially improve the results.
Therefore, we believe that the presented results can be further improved, and encourage future research to explore more powerful mechanisms to provide more accurate proportions of the object of interest.

\section*{Acknowledgements}

This work is supported by the National Science and Engineering Research Council of Canada (NSERC), Fonds de recherche du Québec (FRQNT), and Prompt Quebec. We also thank Calcul Quebec and Compute Canada.

{\small
\bibliographystyle{ieee_fullname}
\bibliography{references}

\begin{thebibliography}{10}\itemsep=-1pt

\bibitem{boudiaffew2021}
Malik Boudiaf, Hoel Kervadec, Ziko~Imtiaz Masud, Pablo Piantanida, Ismail
  Ben~Ayed, and Jose Dolz.
\newblock Few-shot segmentation without meta-learning: A good transductive
  inference is all you need?
\newblock In {\em Proceedings of the IEEE/CVF Conference on Computer Vision and
  Pattern Recognition}, pages 13979--13988, 2021.

\bibitem{boudiaf2020information}
Malik Boudiaf, Imtiaz Ziko, J{\'e}r{\^o}me Rony, Jos{\'e} Dolz, Pablo
  Piantanida, and Ismail Ben~Ayed.
\newblock Information maximization for few-shot learning.
\newblock {\em Advances in Neural Information Processing Systems},
  33:2445--2457, 2020.

\bibitem{castro2018end}
Francisco~M Castro, Manuel~J Mar{\'\i}n-Jim{\'e}nez, Nicol{\'a}s Guil, Cordelia
  Schmid, and Karteek Alahari.
\newblock End-to-end incremental learning.
\newblock In {\em Proceedings of the European conference on computer vision
  (ECCV)}, pages 233--248, 2018.

\bibitem{cermelli2020modeling}
Fabio Cermelli, Massimiliano Mancini, Samuel~Rota Bulo, Elisa Ricci, and
  Barbara Caputo.
\newblock Modeling the background for incremental learning in semantic
  segmentation.
\newblock In {\em Proceedings of the IEEE/CVF Conference on Computer Vision and
  Pattern Recognition}, pages 9233--9242, 2020.

\bibitem{dhillon2019baseline}
Guneet~Singh Dhillon, Pratik Chaudhari, Avinash Ravichandran, and Stefano
  Soatto.
\newblock A baseline for few-shot image classification.
\newblock In {\em International Conference on Learning Representations}, 2019.

\bibitem{dong2022federated}
Jiahua Dong, Lixu Wang, Zhen Fang, Gan Sun, Shichao Xu, Xiao Wang, and Qi Zhu.
\newblock Federated class-incremental learning.
\newblock In {\em Proceedings of the IEEE/CVF Conference on Computer Vision and
  Pattern Recognition}, pages 10164--10173, 2022.

\bibitem{dong2018few}
Nanqing Dong and Eric~P Xing.
\newblock Few-shot semantic segmentation with prototype learning.
\newblock In {\em BMVC}, volume~3, 2018.

\bibitem{everingham2015pascal}
Mark Everingham, SM Eslami, Luc Van~Gool, Christopher~KI Williams, John Winn,
  and Andrew Zisserman.
\newblock The pascal visual object classes challenge: A retrospective.
\newblock {\em International journal of computer vision}, 111(1):98--136, 2015.

\bibitem{finn2017model}
Chelsea Finn, Pieter Abbeel, and Sergey Levine.
\newblock Model-agnostic meta-learning for fast adaptation of deep networks.
\newblock In {\em International conference on machine learning}, pages
  1126--1135. PMLR, 2017.

\bibitem{hariharan2014simultaneous}
Bharath Hariharan, Pablo Arbel{\'a}ez, Ross Girshick, and Jitendra Malik.
\newblock Simultaneous detection and segmentation.
\newblock In {\em European conference on computer vision}, pages 297--312.
  Springer, 2014.

\bibitem{he2016deep}
Kaiming He, Xiangyu Zhang, Shaoqing Ren, and Jian Sun.
\newblock Deep residual learning for image recognition.
\newblock In {\em Proceedings of the IEEE conference on computer vision and
  pattern recognition}, pages 770--778, 2016.

\bibitem{hinton2015distilling}
Geoffrey Hinton, Oriol Vinyals, and Jeffrey Dean.
\newblock Distilling the knowledge in a neural network.
\newblock In {\em NIPS Deep Learning and Representation Learning Workshop},
  2015.

\bibitem{jabi2019deep}
Mohammed Jabi, Marco Pedersoli, Amar Mitiche, and Ismail~Ben Ayed.
\newblock Deep clustering: On the link between discriminative models and
  k-means.
\newblock {\em IEEE transactions on pattern analysis and machine intelligence},
  43(6):1887--1896, 2019.

\bibitem{lang2022learning}
Chunbo Lang, Gong Cheng, Binfei Tu, and Junwei Han.
\newblock Learning what not to segment: A new perspective on few-shot
  segmentation.
\newblock In {\em Proceedings of the IEEE/CVF Conference on Computer Vision and
  Pattern Recognition (CVPR)}, pages 8057--8067, 2022.

\bibitem{li2021adaptive}
Gen Li, Varun Jampani, Laura Sevilla-Lara, Deqing Sun, Jonghyun Kim, and
  Joongkyu Kim.
\newblock Adaptive prototype learning and allocation for few-shot segmentation.
\newblock In {\em Proceedings of the IEEE/CVF Conference on Computer Vision and
  Pattern Recognition}, pages 8334--8343, 2021.

\bibitem{li2017learning}
Zhizhong Li and Derek Hoiem.
\newblock Learning without forgetting.
\newblock {\em IEEE transactions on pattern analysis and machine intelligence},
  40(12):2935--2947, 2017.

\bibitem{lin2014microsoft}
Tsung-Yi Lin, Michael Maire, Serge Belongie, James Hays, Pietro Perona, Deva
  Ramanan, Piotr Doll{\'a}r, and C~Lawrence Zitnick.
\newblock Microsoft coco: Common objects in context.
\newblock In {\em European conference on computer vision}, pages 740--755.
  Springer, 2014.

\bibitem{linsker1988self}
Ralph Linsker.
\newblock Self-organization in a perceptual network.
\newblock {\em Computer}, 21(3):105--117, 1988.

\bibitem{liu2018learning}
Yanbin Liu, Juho Lee, Minseop Park, Saehoon Kim, Eunho Yang, Sung~Ju Hwang, and
  Yi Yang.
\newblock Learning to propagate labels: Transductive propagation network for
  few-shot learning.
\newblock In {\em International Conference on Learning Representations}, 2019.

\bibitem{liu2022intermediate}
Yuanwei Liu, Nian Liu, Xiwen Yao, and Junwei Han.
\newblock Intermediate prototype mining transformer for few-shot semantic
  segmentation.
\newblock In {\em Neural Information Processing Systems (NeurIPS)}, 2022.

\bibitem{liu2020part}
Yongfei Liu, Xiangyi Zhang, Songyang Zhang, and Xuming He.
\newblock Part-aware prototype network for few-shot semantic segmentation.
\newblock In {\em European Conference on Computer Vision}, pages 142--158.
  Springer, 2020.

\bibitem{lu2021simpler}
Zhihe Lu, Sen He, Xiatian Zhu, Li Zhang, Yi-Zhe Song, and Tao Xiang.
\newblock Simpler is better: Few-shot semantic segmentation with classifier
  weight transformer.
\newblock In {\em Proceedings of the IEEE/CVF International Conference on
  Computer Vision}, pages 8741--8750, 2021.

\bibitem{rakelly2018conditional}
Kate Rakelly, Evan Shelhamer, Trevor Darrell, Alyosha Efros, and Sergey Levine.
\newblock Conditional networks for few-shot semantic segmentation.
\newblock In {\em International Conference on Learning Representations (ICLR)
  Workshop}, 2018.

\bibitem{ravi2016optimization}
Sachin Ravi and Hugo Larochelle.
\newblock Optimization as a model for few-shot learning.
\newblock In {\em International Conference on Learning Representations (ICLR)},
  2017.

\bibitem{shaban2017oneshot}
Amirreza Shaban, Shray Bansal, Liu Zhen, Irfan Essa, and Byron Boots.
\newblock One-shot learning for semantic segmentation.
\newblock In {\em Proceedings of the British Machine Vision Conference (BMVC)},
  pages 167.1--167.13. BMVA Press, September 2017.

\bibitem{shannon1948mathematical}
Claude~Elwood Shannon.
\newblock A mathematical theory of communication.
\newblock {\em The Bell system technical journal}, 27(3):379--423, 1948.

\bibitem{siam2019amp}
Mennatullah Siam, Boris~N Oreshkin, and Martin Jagersand.
\newblock Amp: Adaptive masked proxies for few-shot segmentation.
\newblock In {\em Proceedings of the IEEE/CVF International Conference on
  Computer Vision}, pages 5249--5258, 2019.

\bibitem{sun2022singular}
Yanpeng Sun, Qiang Chen, Xiangyu He, Jian Wang, Haocheng Feng, Junyu Han, Errui
  Ding, Jian Cheng, Zechao Li, and Jingdong Wang.
\newblock Singular value fine-tuning: Few-shot segmentation requires
  few-parameters fine-tuning.
\newblock In {\em NeurIPS}, 2022.

\bibitem{tian2022generalized}
Zhuotao Tian, Xin Lai, Li Jiang, Shu Liu, Michelle Shu, Hengshuang Zhao, and
  Jiaya Jia.
\newblock Generalized few-shot semantic segmentation.
\newblock In {\em Proceedings of the IEEE/CVF Conference on Computer Vision and
  Pattern Recognition}, pages 11563--11572, 2022.

\bibitem{tian2020prior}
Zhuotao Tian, Hengshuang Zhao, Michelle Shu, Zhicheng Yang, Ruiyu Li, and Jiaya
  Jia.
\newblock Prior guided feature enrichment network for few-shot segmentation.
\newblock {\em IEEE transactions on pattern analysis and machine intelligence},
  2020.

\bibitem{veilleux2021realistic}
Olivier Veilleux, Malik Boudiaf, Pablo Piantanida, and Ismail Ben~Ayed.
\newblock Realistic evaluation of transductive few-shot learning.
\newblock {\em Advances in Neural Information Processing Systems},
  34:9290--9302, 2021.

\bibitem{wang2020few}
Haochen Wang, Xudong Zhang, Yutao Hu, Yandan Yang, Xianbin Cao, and Xiantong
  Zhen.
\newblock Few-shot semantic segmentation with democratic attention networks.
\newblock In {\em European Conference on Computer Vision}, pages 730--746.
  Springer, 2020.

\bibitem{wang2019panet}
Kaixin Wang, Jun~Hao Liew, Yingtian Zou, Daquan Zhou, and Jiashi Feng.
\newblock Panet: Few-shot image semantic segmentation with prototype alignment.
\newblock In {\em Proceedings of the IEEE/CVF International Conference on
  Computer Vision}, pages 9197--9206, 2019.

\bibitem{yang2020prototype}
Boyu Yang, Chang Liu, Bohao Li, Jianbin Jiao, and Qixiang Ye.
\newblock Prototype mixture models for few-shot semantic segmentation.
\newblock In {\em European Conference on Computer Vision}, pages 763--778.
  Springer, 2020.

\bibitem{ye2021learning}
Han-Jia Ye, Hexiang Hu, and De-Chuan Zhan.
\newblock Learning adaptive classifiers synthesis for generalized few-shot
  learning.
\newblock {\em International Journal of Computer Vision}, 129:1930--1953, 2021.

\bibitem{zhang2021self}
Bingfeng Zhang, Jimin Xiao, and Terry Qin.
\newblock Self-guided and cross-guided learning for few-shot segmentation.
\newblock In {\em Proceedings of the IEEE/CVF Conference on Computer Vision and
  Pattern Recognition}, pages 8312--8321, 2021.

\bibitem{zhang2019canet}
Chi Zhang, Guosheng Lin, Fayao Liu, Rui Yao, and Chunhua Shen.
\newblock Canet: Class-agnostic segmentation networks with iterative refinement
  and attentive few-shot learning.
\newblock In {\em Proceedings of the IEEE/CVF Conference on Computer Vision and
  Pattern Recognition}, pages 5217--5226, 2019.

\bibitem{zhang2021few}
Gengwei Zhang, Guoliang Kang, Yi Yang, and Yunchao Wei.
\newblock Few-shot segmentation via cycle-consistent transformer.
\newblock {\em Advances in Neural Information Processing Systems},
  34:21984--21996, 2021.

\bibitem{zhao2017pyramid}
Hengshuang Zhao, Jianping Shi, Xiaojuan Qi, Xiaogang Wang, and Jiaya Jia.
\newblock Pyramid scene parsing network.
\newblock In {\em Proceedings of the IEEE conference on computer vision and
  pattern recognition}, pages 2881--2890, 2017.

\bibitem{ziko2020laplacian}
Imtiaz Ziko, Jose Dolz, Eric Granger, and Ismail~Ben Ayed.
\newblock Laplacian regularized few-shot learning.
\newblock In {\em International conference on machine learning}, pages
  11660--11670. PMLR, 2020.

\end{thebibliography}
}

\clearpage
\appendix

\section{Datasets} \label{app:datasets}

We evaluated our method on two widely-used few-shot segmentation benchmarks: \pascalfivei{} \cite{shaban2017oneshot} and \coco{} \cite{shaban2017oneshot}.
The former is built based on \mbox{PASCAL VOC 2012 \cite{everingham2015pascal}} (containing 20 semantic classes) with additional annotations from SDS \cite{hariharan2014simultaneous}, while the latter is built from \mbox{MS-COCO} \cite{lin2014microsoft} (containing 80 semantic classes).
In both datasets, the classes are split into 4 disjoint subsets, and the experiments are done in a cross-validation manner. For each fold, the set of novel classes is extracted from one of these subsets while the union of the remaining subsets will be the set of base classes. Furthermore, as discussed in the main paper, we have introduced a new scenario, where the number of novel classes is increased, referred to as \pascalteni. The semantic classes in each fold of \pascalteni{} are detailed in \cref{table:pascal10i-classes}.

\begin{table*}[bp]
    \centering
    \begin{tabular}{c|c|c}
         & Novel classes & Base classes \\
        \midrule
        \multirow{2}{*}{\mbox{PASCAL-$10^0$}} & aeroplane, bicycle, bird, boat, bottle, & diningtable, dog, horse, motorbike, person, \\
        & bus, car, cat, chair, cow & potted plant, sheep, sofa, train, tv/monitor \\
        \midrule
        \multirow{2}{*}{\mbox{PASCAL-$10^1$}} & diningtable, dog, horse, motorbike, person, & aeroplane, bicycle, bird, boat, bottle, \\
        & potted plant, sheep, sofa, train, tv/monitor & bus, car, cat, chair, cow \\
    \end{tabular}
    \caption{\textbf{Semantic classes in each fold of \pascalteni.} In this benchmark, each fold contains 10 novel classes and hence introduces more difficulties in the generalized few-shot segmentation scenario.}
    \label{table:pascal10i-classes}
\end{table*}

\section{Ablation on the number of iterations} \label{app:num-iter}

\begin{figure}
    \centering
    \includegraphics[width=\linewidth]{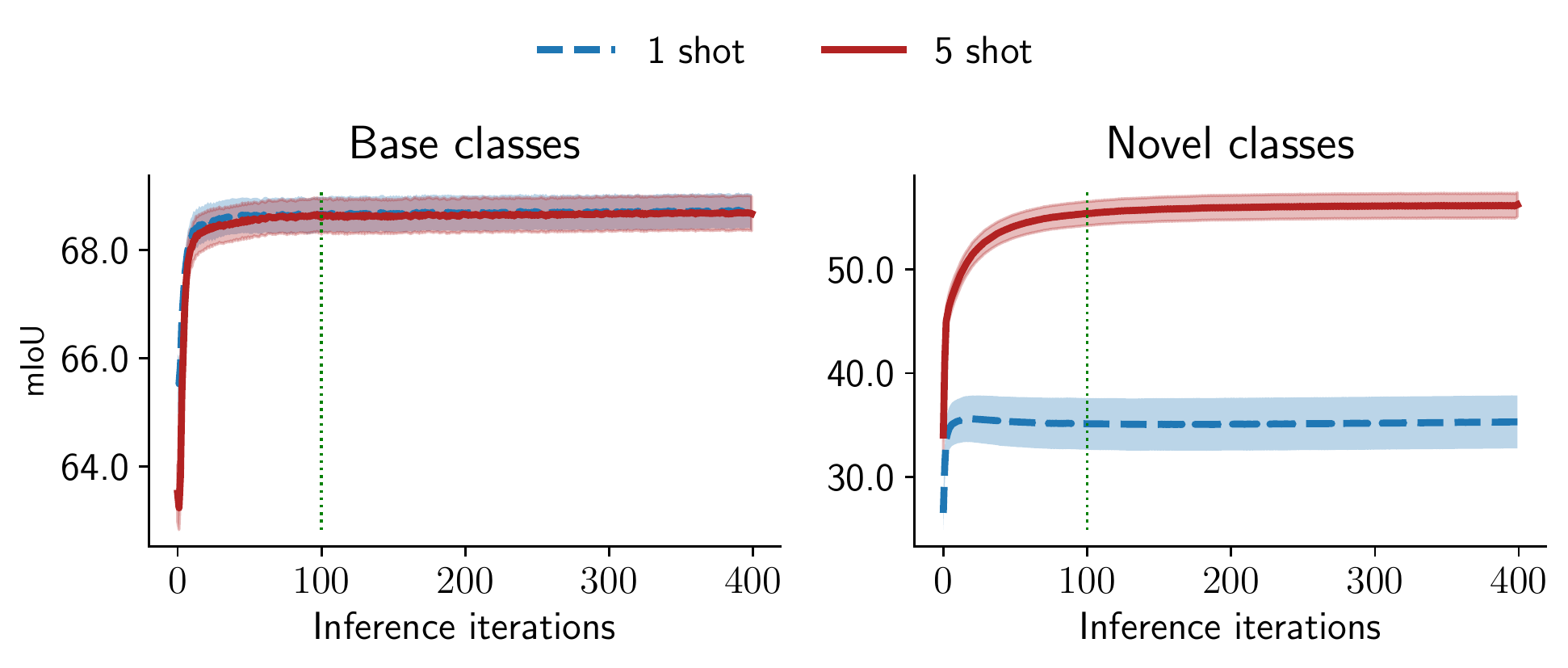}
    \caption{\textbf{Performance of the method as the number of iterations in the adaptation phase increases.} Metrics improve rapidly and first and the improvements slow down as the model is further optimized. The dotted green line indicates our choice for the number of iterations in the main manuscript. Results are provided for \pascalfivei.}
    \label{fig:ablation-num-iter}
\end{figure}

In our empirical validation, the proposed loss function, $\mathcal{L}_{\ourmethod}$, is optimized for a fixed number of iterations ($n=100$), \textit{which was chosen arbitrarily.}  As demonstrated in \cref{fig:ablation-num-iter} and \cref{table:ablation-num-iter} the metrics reach high values using only a few iterations. This finding shows that we can speed up the adaptation further by reducing the number of iterations while keeping the performance relatively intact. Also, continuing the adaptation for longer does not hurt the performance and the metrics stay almost the same. Please note that, as stated, the number of iterations in all the experiments in the main manuscript was set to 100 arbitrarily, disregarding the findings of this section.

\begin{table}[htb]
    \centering
    \resizebox{1.0\linewidth}{!}{
    \begin{tabular}{lcccccccc}
	    \toprule
	    && \multicolumn{3}{c}{1-Shot} && \multicolumn{3}{c}{5-Shot} \\
	    \cmidrule{3-9}
	    \# iterations && Base & Novel & \newmetric && Base & Novel & \newmetric \\
	    \midrule
	    10  && 70.15 & 35.38 & 52.77 && 69.92 & 48.51 & 59.22 \\
	    50  && 70.79 & 35.33 & 53.06 && 70.63 & 54.15 & 62.39 \\
	    \rowcolor{lightgray!40}
	    100 && 70.89 & 35.11 & 53.00 && 70.85 & 55.31 & 63.08 \\
	    200 && 70.87 & 35.10 & 52.99 && 70.81 & 56.00 & 63.41 \\
	    300 && 70.91 & 35.18 & 53.05 && 70.83 & 56.19 & 63.51 \\
	    400 && 70.88 & 35.30 & 53.09 && 70.85 & 56.22 & 63.54 \\
	    \bottomrule
    \end{tabular}
    }
    \caption{\textbf{Precise values of the performance metrics, at selected points on the plots in \cref{fig:ablation-num-iter}.} The shaded row indicates our choice for the number of iterations reported in the main manuscript. Results are provided for \pascalfivei.}
    \label{table:ablation-num-iter}
\end{table}

\section{Detailed results}
The evaluation of our approach is performed in a cross-validation manner. In particular, there exist 4 folds for each of \pascalfivei{} and \coco{} benchmarks, and 2 folds for \pascalteni. In the main manuscript, the reported results are obtained by averaging over all the folds in these benchmarks. \Cref{table:detailed} shows the performance of our model on each fold individually.

\begin{table}[tbh]
	\centering
	\resizebox{1.0\linewidth}{!}{
	\begin{tabular}{lccccccccc}
	    \toprule
	    &&& \multicolumn{3}{c}{1-Shot} && \multicolumn{3}{c}{5-Shot} \\
	    \cmidrule{4-10}
	    Benchmark & Fold && Base & Novel & \newmetric && Base & Novel & \newmetric \\
	    \midrule
	    \multirow{5}{*}{\pascalfivei{}}
	    & 0 &&  71.33 & 29.36 & 50.35 && 71.06 & 53.72 & 62.39 \\
	    & 1 &&  69.54 & 46.72 & 58.13 && 69.63 & 63.33 & 66.48 \\
	    & 2 &&  69.10 & 27.07 & 48.09 && 69.12 & 54.01 & 61.57 \\
	    & 3 &&  73.60 & 37.30 & 55.45 && 73.60 & 50.19 & 61.90 \\
	    \rowcolor{lightgray!40}
	    \cellcolor{white} & mean &&  70.89 & 35.11 & 53.00 && 70.85 & 55.31 & 63.08 \\
	    \midrule
	    \multirow{5}{*}{\coco{}}
	    & 0 && 49.01 & 15.89 & 32.45 && 48.90 & 24.86 & 36.88 \\
	    & 1 && 46.83 & 19.50 & 33.17 && 47.10 & 33.94 & 40.52 \\
	    & 2 && 48.82 & 16.93 & 32.88 && 49.12 & 27.15 & 38.14 \\
	    & 3 && 48.45 & 16.57 & 32.51 && 48.37 & 28.95 & 38.66 \\
	    \rowcolor{lightgray!40}
	    \cellcolor{white} & mean && 48.28 & 17.22 & 32.75 && 48.37 & 28.73 & 38.55 \\
	    \midrule
	    \multirow{3}{*}{\pascalteni{}}
	    & 0 && 68.69 & 34.40 & 51.55 && 68.49 & 55.94 & 62.22 \\
	    & 1 && 71.83 & 28.17 & 50.00 && 72.00 & 47.84 & 59.92 \\
	    \rowcolor{lightgray!40}
	    \cellcolor{white} & mean && 70.26 & 31.29 & 50.77 && 70.25 & 51.89 & 61.07 \\
	    \bottomrule
	\end{tabular}
	}
	\caption{\textbf{Detailed results for each fold.} For each of the benchmarks, the performance of our method is presented for all the folds.}
	\label{table:detailed}
\end{table}

\section{Harmonic mean}
Following \cite{ye2021learning}, we provide in \cref{table:h-mean} the harmonic mean score, referred to as \textit{H-Mean}, of CAPL \cite{tian2022generalized}, BAM \cite{lang2022learning}, and our method for reference. Using this metric increases the overall performance gap between our method and existing approaches.

\begin{table}[ht!]
	\centering
	\resizebox{1.0\linewidth}{!}
	{
	\begin{tabular}{@{}l@{\hskip 3px}lcccccccc@{}}
	    \toprule
	    && \multicolumn{7}{c}{\textbf{\pascalfivei{}}}\\
	     && \multicolumn{3}{c}{1-Shot} && \multicolumn{3}{c}{5-Shot} \\
	    \cmidrule{3-9}
	    Method && Base & Novel & H-Mean && Base & Novel & H-Mean \\
	    \midrule
	    CAPL \cite{tian2022generalized} && 64.80 & 17.46 & 27.51 &&    65.43 & 24.43 & 35.58 \\
	    BAM \cite{lang2022learning} && \textbf{71.60} & 27.49 & 39.73 &&    \textbf{71.60} & 28.96 & 41.24 \\
	    \ourmethod{} && 70.89 & \textbf{35.11} & \textbf{46.96} && 70.85 & \textbf{55.31} & \textbf{62.12} \\
	    \midrule
	    & \\
	    && \multicolumn{7}{c}{\textbf{\coco{}}} \\
	    \cmidrule{3-9}
	    && Base & Novel & H-Mean && Base & Novel & H-Mean \\
	    \midrule
	    CAPL \cite{tian2022generalized} && 43.21 & 7.21 & 12.36 &&   43.71 & 11.00 & 17.58 \\
	    BAM \cite{lang2022learning} && \textbf{49.84} & 14.16 & 22.05 && \textbf{49.85} & 16.63 & 24.94 \\
	    \ourmethod{} && 48.28 & \textbf{17.22} & \textbf{25.39} && 48.37 & \textbf{28.73} & \textbf{36.05} \\
	    \bottomrule
	\end{tabular}
	}
	\caption{\textbf{Quantitative evaluation on \pascalfivei{} and \coco{} compared to GFSS methods, using harmonic mean as the overall score.} 
	}
	\label{table:h-mean}
\end{table}

\section{Including the background in the base score} \label{app:including-bg}

CAPL \cite{tian2022generalized} takes into account the background IoU, which is generally higher than the IoU of base classes, when computing the \textit{Base} metric. We, the same as \cite{lang2022learning}, believe that since background does not represent an object of interest, the model's performance on this class should not be considered. Nevertheless, including the background IoU in the metrics leads to marginal performance differences that are consistent across all methods. In \cref{table:base-incl-bg}, for GFSS methods, the background IoU is included in the \textit{Base} metric, reframing it as \textit{Base w/ bg} to avoid confusion.

\begin{table}[htb]
	\centering
	\resizebox{1.0\linewidth}{!}
	{
	\begin{tabular}{@{}l@{\hskip 3px}lcccccccc@{}}
	    \toprule
	     && \multicolumn{3}{c}{1-Shot} && \multicolumn{3}{c}{5-Shot} \\
	    \cmidrule{3-9}
	    Method && Base w/ bg & Novel & \newmetric && Base w/ bg & Novel & \newmetric \\
	    \midrule
	    CAPL \cite{tian2022generalized} && 66.37 & 17.46 & 41.92 && 66.95 & 24.43 & 45.69 \\
	    BAM \cite{lang2022learning} && 72.00 & 27.49 & 49.75 && \textbf{72.36} & 28.96 & 50.66 \\
	    \ourmethod{} && \textbf{72.04} & \textbf{35.11} & \textbf{53.58} && 72.12 & \textbf{55.31} & \textbf{63.72} \\
	    \bottomrule
	\end{tabular}
	}
	\caption{\textbf{Quantitative evaluation on \pascalfivei{} compared to GFSS methods, including the background performance in the metrics.} 
	}
	\label{table:base-incl-bg}
\end{table}

\section{Practical setting: employing the whole training dataset} \label{app:practical}

As discussed in the main manuscript, CAPL \cite{tian2022generalized} and BAM \cite{lang2022learning} filter out training images that contain novel classes. This procedure is impractical in real-world scenarios since it needs a training set in which novel classes are labeled, undermining the goal of few-shot learning, \ie, having only a few labeled examples of the novel classes. Recent empirical evidence \cite{sun2022singular} has shown that such additional step can lead to performance gain on novel classes. In \cref{table:practical-setting}, we have changed the training procedure of CAPL and BAM and avoided removing images containing novel classes from training. More specifically, 
the potential objects from novel classes are labeled as \textit{background} during training.

\begin{table}[htb]
	\centering
	\resizebox{1.0\linewidth}{!}
	{
	\begin{tabular}{@{}l@{\hskip 3px}lcccccccc@{}}
	    \toprule
	     && \multicolumn{3}{c}{1-Shot} && \multicolumn{3}{c}{5-Shot} \\
	    \cmidrule{3-9}
	    Method && Base & Novel & \newmetric && Base & Novel & \newmetric \\
	    \midrule
	    CAPL \cite{tian2022generalized} && 71.59 & 12.69 & 42.14 && \textbf{71.71} & 19.58 & 45.65 \\
	    BAM \cite{lang2022learning} && \textbf{71.61} & 19.35 & 45.48 && 71.66 & 26.33 & 49.00 \\
	    \ourmethod{} && 70.89 & \textbf{35.11} & \textbf{53.00} && 70.85 & \textbf{55.31} & \textbf{63.08} \\
	    \bottomrule
	\end{tabular}
	}
	\caption{\textbf{Quantitative evaluation on \pascalfivei{} compared to GFSS methods, in the experimental setting in which the whole training dataset is employed.} In this setting, images containing novel classes are not removed from the training process. 
	Confirming the findings in \cite{sun2022singular}, this procedure enhances the performance on novel classes. It is also worth noting that although \textit{Novel} score has been decreased for CAPL, its \textit{Base} score has been considerably increased.
	}
	\label{table:practical-setting}
\end{table}

\section{Adaptation of BAM to multi-class GFSS} \label{app:bam-adaptation}

The results reported in \cite{lang2022learning} for the GFSS task are based on an evaluation protocol in which only one novel class can be recognized in a query image. Indeed, the \textit{meta-learner} from this method can only provide binary (\ie, \textit{ background vs foreground}) predictions and is not practical in the setting where multiple novel classes are to be predicted at the same time. To be able to incorporate BAM in our empirical validation, we had to adapt it so that it can predict multiple novel classes simultaneously. These modifications are detailed in what follows. 
First, instead of selecting $K$ support samples of a novel class $c$ and asking the model to segment class $c$ in a query image, we form $\Nn$ different support sets, $\support_i$, one for each $i \in \Cn$. Recall that $\Cn$ is the set of novel classes and that $\support_i$ contains $K$ samples labeled for each novel class $i$. Second, we run BAM $\Nn$ times and, in each inference, we give the same query image alongside $\support_i$, resulting in a foreground probability map for each class $i$, called $\mathbf{m}_i$. Then, we need to create a single mask containing all the novel class predictions to further use it in BAM's fusion mechanism. To do this, we create an aggregated novel map, $\mathbf{a}$, which is formed based on the resulted $\Nn$ maps, in such a way that for each pixel $j$:

\begin{equation}
    \phantom{.}\mathbf{a}(j) = \argmax_{i \in \Cn} \; \mathbf{m}_i(j).
\end{equation}

We also form $\pp_\mathbf{a}$ to preserve the probability of the selected indices, which will later be compared to the predefined threshold $\tau$ introduced in \cite{lang2022learning}:

\begin{equation}
    \phantom{.}
    \pp_\mathbf{a}(j) = \max_{i \in \Cn} \; \mathbf{m}_i(j).
\end{equation}

Then, $\mathbf{a}$ and $\pp_\mathbf{a}$ alongside the base map predicted by BAM's \textit{base-learner} for the query, $\hat{\mathbf{m}}_b$, are used to perform the fusion procedure following \cite{lang2022learning}. More specifically, the final prediction is formulated as

\begin{numcases}{\hat{\mathbf{m}}_g(j) =}
    \mathbf{a}(j) & $\pp_\mathbf{a}(j) > \tau$, \nonumber \\
    \hat{\mathbf{m}}_b(j) & $\pp_\mathbf{a}(j) \le \tau$ \; and \; $\hat{\mathbf{m}}_b(j) \ne 0$, \\
    0 & otherwise. \nonumber
\end{numcases}

This change definitely slows the inference by an order of magnitude, but this is inevitable because of the nature of meta-learning few-shot segmentation, which needs to be accommodated to produce multi-class semantic maps, as they are tailored to binary maps.

\section{Visual examples} \label{app:visual}

\begin{figure*}[tb]
    \centering
    \begin{subfigure}[t]{\linewidth}
        \centering
        \includegraphics[width=\linewidth]{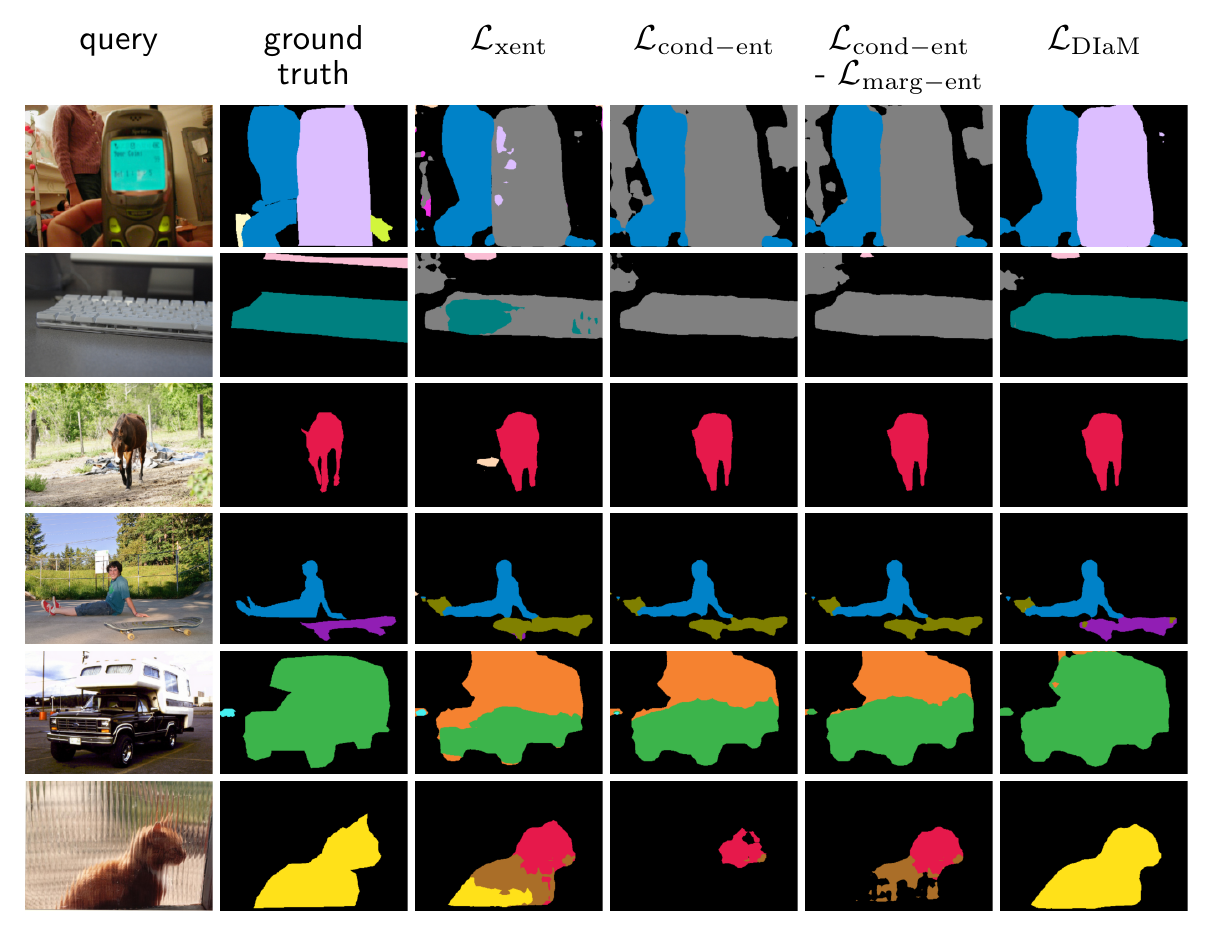}
    \end{subfigure}
    \begin{subfigure}[t]{\linewidth}
        \centering
        \includegraphics[width=1\linewidth]{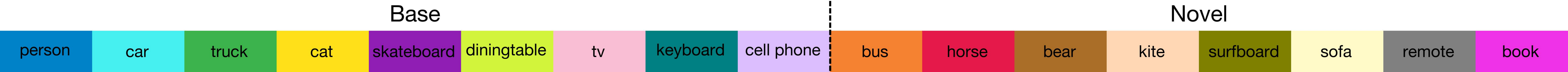}
    \end{subfigure}
    \caption{\textbf{Qualitative results of different terms of \ourmethod's loss function on \coco{}.} A single support set, containing the following novel classes is used for predicting every query image:
    \textit{bicycle}, \textit{bus}, \textit{traffic light}, \textit{bench}, \textit{horse}, \textit{bear}, \textit{umbrella}, \textit{frisbee}, \textit{kite}, \textit{surfboard}, \textit{cup}, \textit{bowl}, \textit{orange}, \textit{pizza}, \textit{sofa}, \textit{toilet}, \textit{remote}, \textit{oven}, \textit{book}, \textit{teddy bear}. 
    Query images can contain any classes and every one of them is to be recognized. From the left, the first two columns show the query image and the ground truth, and the following columns display predictions of models using different loss functions. Results are on \coco{} under the 5-shot setting.}
    \label{fig:coco-visual}
\end{figure*}

In the main manuscript, we presented qualitative results on \pascalfivei{} using different versions of our loss function. We observed that in the absence of the knowledge distillation term, the model misclassifies some of the previously learned base classes as novel ones. \Cref{fig:coco-visual} shows similar results on \coco, where the same trend is observed. For instance, in the first two rows, base classes \textit{cell phone} and \textit{keyboard} are mistakenly classified as the novel class \textit{remote}. Note that this problem is fixed when the knowledge distillation term is added to the loss function.

\end{document}